\documentclass[conference]{IEEEtran}
\IEEEoverridecommandlockouts
\usepackage{cite}
\usepackage{amsmath,amssymb,amsfonts}
\usepackage{algorithmic}
\usepackage{graphicx}
\usepackage{textcomp}
\usepackage{xcolor}

\usepackage[normalem]{ulem}
\usepackage{makecell}
\usepackage{mathtools}
\usepackage{multirow}
\usepackage[symbol]{footmisc}

\newcommand{\etal}{\textit{et al}.}

\newcommand{\adasted}{ST-AED}

\def\BibTeX{{\rm B\kern-.05em{\sc i\kern-.025em b}\kern-.08em
    T\kern-.1667em\lower.7ex\hbox{E}\kern-.125emX}}
\begin{document}

\title{Angle Range and Identity Similarity Enhanced Gaze and Head Redirection based on Synthetic data
}

\author{\IEEEauthorblockN{
Jiawei Qin
\thanks{This work was conducted during the first author's internship at CyberAgent, Inc.}
}
\IEEEauthorblockA{
\textit{The University of Tokyo}\\
Tokyo, Japan \\
jqin@iis.u-tokyo.ac.jp}
\and
\IEEEauthorblockN{
Xueting Wang}
\IEEEauthorblockA{
\textit{CyberAgent Inc.}\\
Tokyo, Japan \\ 
wang\_xueting@cyberagent.co.jp}
}

\maketitle

\begin{abstract}
In this paper, we propose a method for improving the angular accuracy and photo-reality of gaze and head redirection in full-face images. 
The problem with current models is that they cannot handle redirection at large angles, and this limitation mainly comes from the lack of training data.
To resolve this problem, we create data augmentation by monocular 3D face reconstruction to extend the head pose and gaze range of the real data, which allows the model to handle a wider redirection range.
In addition to the main focus on data augmentation, we also propose a framework with better image quality and identity preservation of unseen subjects even training with synthetic data.
Experiments show that our method significantly improves redirection performance in terms of redirection angular accuracy while maintaining high image quality, especially when redirecting to large angles.
\end{abstract}



\section{Introduction}

In recent years, gaze-related research has gained widespread interest due to its significant role in many applications such as human-computer interaction and human behavior analysis. 
For example, gaze estimation~\cite{gaze_survey_ghosh2021automatic} makes great progress with the development of deep learning. 
Redirecting the gaze and head of a given image becomes an important topic due to the growing demand for semi-supervised or unsupervised training for gaze estimation~\cite{gaze_survey_ghosh2021automatic, sted}, and it is also can be applied to diverse digital face synthesis.
There are many works focused on redirecting gaze based on image-warping~\cite{ganin2016deepwarp}, graphical models~\cite{wood2018gazedirector}, or generative adversarial network (GAN)~\cite{he2019GAN_eye}. 
While these methods did not consider the full-face input including the head pose factor. 
Recently, another learning-based face synthesis, ST-ED~\cite{sted}, has shown the simultaneous redirection of gaze and head for full-face images.

\begin{figure}[t]
\centering
    \includegraphics[width=0.97\linewidth]{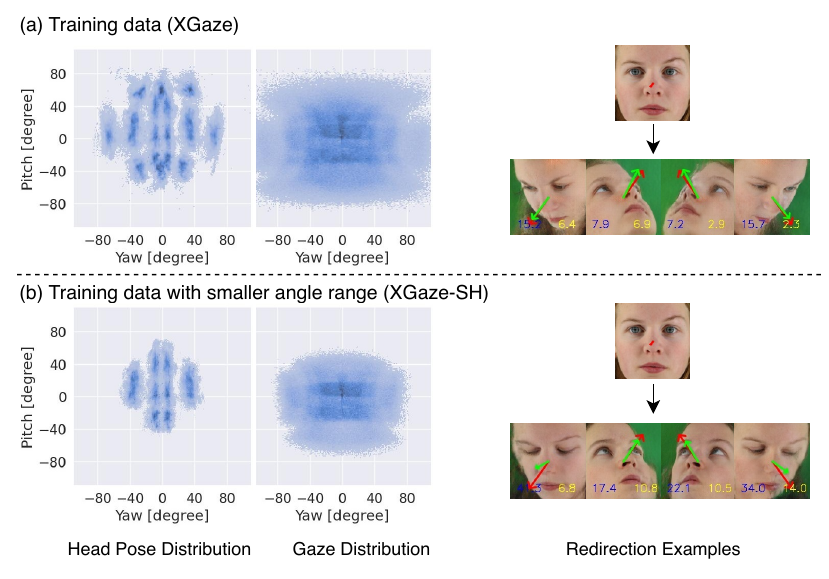}
    \caption{Illustration of the angle range limitation of current learning-based gaze and head redirection. 
    For datasets with a large range (top row) and a small range (bottom row), the head pose and gaze distributions along with examples of the redirected images by these training data are shown from left to right. 
    When training with a smaller angle range, the prior work ST-ED~\cite{sted} fails to redirect to a large target angle (bottom), as compared to the wider training data (top).
    }
    \label{fig:teaser}
\end{figure}

Learning-based gaze and head redirection requires gaze and head pose labels during training, and collecting comprehensive training data is a large challenge in gaze redirection.
Specifically, while head pose labels are relatively easy to get, sophisticated devices are required to obtain accurate gaze labels.
As illustrated in Fig.~\ref{fig:teaser}, we experimentally showed that the current state-of-the-art model, ST-ED~\cite{sted}, cannot precisely redirect to a large angle that is out of the range of the training dataset.
Previous in-the-wild datasets~\cite{mpiifacegaze_zhang2017s, gazecapture} have successfully captured diverse environments with large subject scales as some examples are shown in the upper part of Fig.~\ref{fig:aug_samples}, but the gaze range is naturally limited by the devices used for collecting data.
On the other hand, large-angle data can be collected in lab-controlled settings~\cite{ETHXGaze_Zhang2020}, but the high cost makes it difficult to reach a large subject scale.
In addition, the identity similarity and photo-reality of the generated image also drop due to the lack of high-quality gaze datasets with a large number of subject scales.

To address the above challenges, we propose a redirection framework including creating synthetic training data with extended angle range and a corresponding redirection model~\adasted~, which learns better using synthetic data with image quality improvement.
For data augmentation, we apply 3D face reconstruction that can preserve the accurate original gaze feature and can be rotated in 3D space without estimation error.
In the redirection model~\adasted~, to reduce the side effect of synthetic data with relatively more noise, we propose to adopt a higher-level loss instead of the pixel loss used in prior works~\cite{FAZE_Park2019ICCV,sted}.
Besides, we leverage a state-of-the-art face recognition model as the identity encoder with an identity loss to further constrain the model to have a better appearance feature extraction.

Our contribution can be summarized as 
\begin{enumerate}
    \item We create augmentation data with a much larger angle range than real data to improve learning-based redirection by 3D face reconstruction.
    \item We propose a gaze and head redirection framework to improve image quality and identity preservation on unseen subjects by a pre-trained identity encoder with a designed identity loss. 
    \item Experimental results proved that the proposal can redirect the gaze and face to a larger target angle range with improved image quality compared with the SOTA redirection work.
\end{enumerate}


\section{Related Works}
\subsection{Face Editing by GANs}
GANs have been widely used in face editing to generate highly realistic images~\cite{gan_survey_kammoun2022generative, karras2018progressive, styleganv1_Karras_CVPR19, interfacegan_shen2020}. 
With the help of high-quality training datasets such as FFHQ~\cite{styleganv1_Karras_CVPR19} and CelebA-HQ~\cite{karras2018progressive}, GAN models can be used to edit facial expressions, age, glasses, pose, and other features. 
Many previous works have focused on training GANs to change the pose of a given image~\cite{ interfacegan_shen2020, styleganv1_Karras_CVPR19}, and most of these methods manipulate the head pose in the latent code space to generate an image with a different head pose. 
However, these GAN-based methods are not directly applicable to redirecting head and gaze, as they are not trained on datasets with gaze labels.

\subsection{Data Augmentation by 3D Face Reconstruction}

The rapid development of monocular 3D face reconstruction has led to improvements in other face-related tasks such as face recognition~\cite{3d_aug_rec_2004, 3d_aug_rec_2017, rotate_render_2020} and gaze estimation~\cite{qin_nv}. 
By rotating the 3D reconstructed face, researchers have been able to create more diverse training data for face recognition and extend the range of poses for gaze estimation. 
The pixel RGB values of the input image are used as the 3D face texture, preserving the original appearance.
The projective-matching process~\cite{qin_nv} allows for the 3D face to be rotated and translated within 3D physical space, allowing for arbitrary manipulation and rendering without the need for additional training.

\subsection{Gaze and Head Redirection}
Gaze redirection was originally for eye-only images by deep warping~\cite{ganin2016deepwarp} method or GAN~\cite{he2019GAN_eye}.
Recently, full-face gaze and head redirection~\cite{sted, cuda_ghr_jindal2021} becomes a more significant topic.
Park~\etal~\cite{FAZE_Park2019ICCV} proposed an encoder-decoder model that disentangles the appearance, head, and gaze features and transforms the source embeddings to target by rotation.
During training, the paired images are used such that the model learns the transformation supervised by the head pose and gaze label.
ST-ED~\cite{sted} focused on improving the image quality performance and took into consideration extraneous factors such as illumination and hue by introducing an unsupervised self-learning pipeline.
Consequently, the model enables redirecting a source image to a target image or any target angle.
Learning-based approaches generally require head pose and gaze labels for training and may depend much on the training data.
Although ST-ED is a state-of-the-art model in gaze and head redirection, its ability to redirect to a very large angle with satisfying image quality is not verified.
Collecting training data with a large subject scale and angle range is not a trivial task, but 3D face reconstruction can create an arbitrary amount of face-rotated images with accurate angles without extra training.
Therefore, we consider 3D face reconstruction for data augmentation synthesis to extend the limited angle range of real data.


\section{Data Augmentation}\label{sec:setting}

\begin{figure}[t]
\centering
    \includegraphics[width=0.90\linewidth]{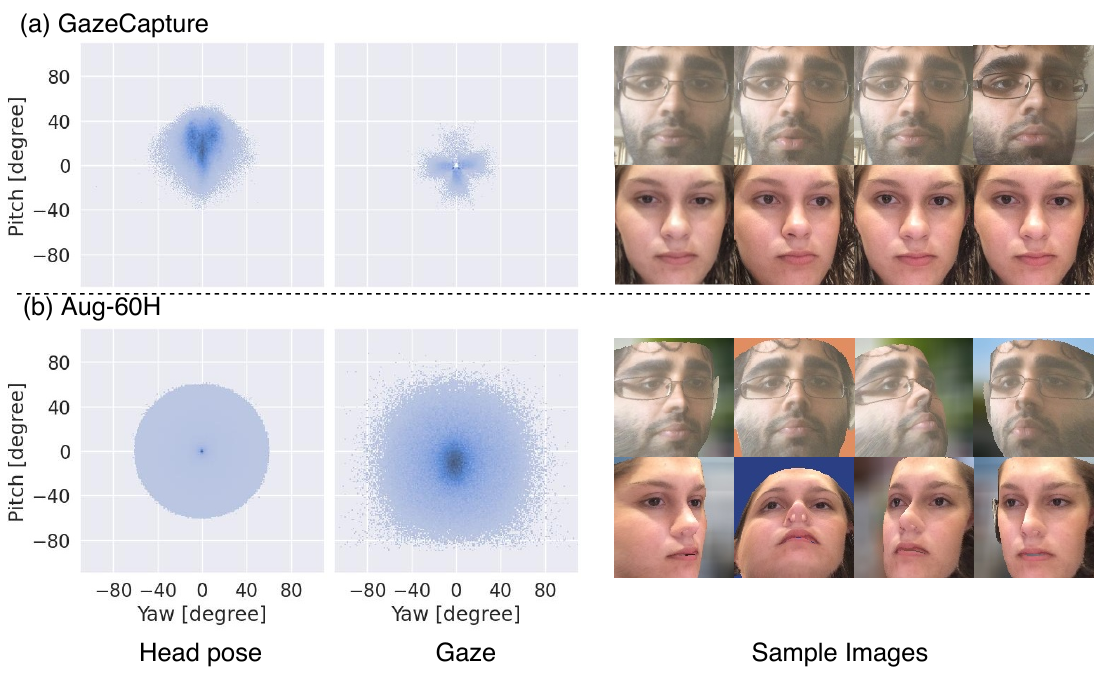}
    \caption{The distribution and samples of original GazeCapture (up) and the augmentation dataset Aug-60H (bottom). }
    \label{fig:aug_samples}
\end{figure}

\subsection{Real Datasets}
\textbf{GazeCapture}~\cite{gazecapture} is collected via crowdsourcing when subjects are using tablets, resulting in a large subject scale with more than 1,400 subjects.  
It totally contains more than 1.6 million images.
However, the angle range is limited as shown in Fig.~\ref{fig:aug_samples}(a).
\textbf{ETH-XGaze}~\cite{ETHXGaze_Zhang2020} put 18 synchronous cameras under different view angles to take pictures, which achieves a very large angle range.
However, it contains only 110 subjects. 
Each subject has around 600 frames, and thus more than 1 million images in total.

We follow the split of ST-ED~\cite{sted} to create GazeCapture Train and GazeCapture Test.
If not specifically mentioned, we abbreviate GazeCapture Train as GazeCapture. 
GazeCapture contains 1,177 subjects with 1,379,083 images, and GazeCapture Test contains 139 subjects with 191,842 images.
We also split the public 80 subjects of the official ETH-XGaze into 70 for training and 10 for testing, denoted as \textbf{XGaze} and \textbf{XGaze Test}, respectively.

\begin{figure*}[t]
\centering
    \includegraphics[width=0.7\linewidth]{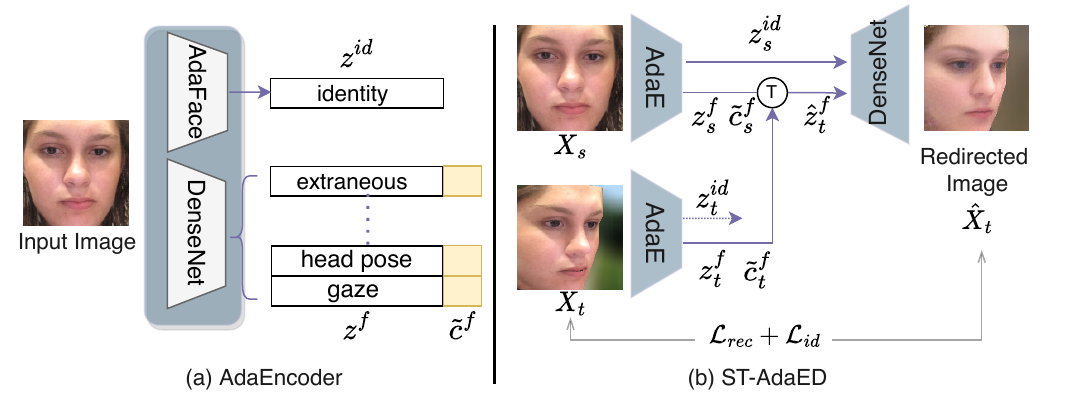}
    \caption{The proposed~\adasted~uses a pre-trained AdaFace network as the identity encoder (left).
    A mixed reconstruction loss $\mathcal{L}_{rec}$ and identity consistency loss $\mathcal{L}_{id}$ are computed between the target image and the redirected image (right).}
    \label{fig:architecture}
\end{figure*}

\subsection{Data Augmentation Creation}\label{aug:method}
We create data augmentation to address the limitation of the above real datasets.
We follow an off-the-shelf 3D face reconstruction pipeline 3DDFA~\cite{3ddfa_cleardusk,3ddfa2_guo2020towards, 3ddfa3_zhu2017face} to do reconstruction and apply the projective-matching~\cite{qin_nv} to finally obtain a 3D face.
Given a target head direction or gaze direction, we can compute a rotation matrix, rotate the face, and render a new image.
And we follow the original setting~\cite{qin_nv} to use random images and random colors as the background for the rendered images.

For augmentation data, we filter out subjects with less than 30 samples in the original GazeCapture, and we randomly sample 30 images from the remained 861 subjects.
Each source image will be augmented to 10 new images, and some augmentation data examples are shown in Fig.~\ref{fig:aug_samples}.
As a result, the augmentation dataset contains 257,470 samples.

In detail, we sample the target head poses from a circle-shaped uniform distribution with a radius of 60\textdegree.
We compute a rotation matrix based on the source head pose and the sampled target head pose (\textit{head-based sampling}). 
Then we rotate the 3D face by this rotation matrix and render new images.
Consequently, the distributions of the original GazeCapture and its augmentation dataset denoted as Aug-60H with `H' representing \textit{head-based}, are shown in Fig.~\ref{fig:aug_samples}.


\section{Redirection Model \adasted}

Through experiment, we found that the generated image by the state-of-the-art baseline ST-ED still cannot realize the faithful identity for unseen subjects, which also results in the dropping of image quality.
Therefore, we propose~\adasted, which tackles these problems by extracting better face identity features for enhancing generated image quality.
The framework of our proposed model~\adasted~is shown in Fig.~\ref{fig:architecture}.
Following ST-ED~\cite{sted}, we use DenseNet to encode the personal-varying factors including head pose, gaze, and other extraneous factors such as lighting and hue.
Each factor is encoded to a pseudo-condition $\tilde{c}$ and embedding $z$, as illustrated in the left side of Fig.~\ref{fig:architecture}.
For head and gaze with the ground-truth label, the transformation of embedding $z$ is computed based on the label and the pseudo-condition $\tilde{c}$.
The redirected image is formulated as $\hat{X}_{t}=D(z^{id}_{s},\hat{z}^{f}_{t})$, where $\hat{z}^{f}_{t}=T(z^{f}_{s}, \hat{c}^{f}_{s}, \hat{c}^{f}_{t})$.

In order to have better generalizability on unseen subjects, we adopt the MS1MV2-pre-trained~\cite{arcface_deng2019} AdaFace~\cite{kim2022adaface} model for encoding identity features, instead of the DenseNet trained from scratch in the baseline ST-ED model.
Besides, since the augmentation data contains relatively more noise such as reconstruction artifacts and random background, we use a mixed loss of MS-SSIM~\cite{ssim_wang2003multiscale} and $\ell1$ loss to avoid over-focusing on the pixel difference inspired from~\cite{Nitzan2020FaceID,mix_l1_zhao}.
Formally, given the target image $X_{t}$ and the generated image $\tilde{X_{t}}$, the reconstruction loss is formulated as 
\begin{equation}
    \mathcal{L}_{rec} = \alpha (1- \textrm{MS-SSIM}(\tilde{X_t}, X_t )) + (1-\alpha) |\tilde{X_t} - X_t|_1.
\end{equation}
While the pre-trained identity encoder is expected to improve the generalization to unseen subjects, 
we further add an identity loss to force the decoded image to have a close identity to the target image, 
$\mathcal{L}_{id} =  1 - \textrm{Sim}(\tilde{X_t}, X_t),$
where the identity similarity is computed using another pre-trained AdaFace face recognition network.
The final loss is formulated as 
\begin{equation}
    \mathcal{L}_{\textrm{total}} = \mathcal{L}_{\textrm{ST-ED}} + \lambda_{id} \mathcal{L}_{id} + \lambda_{rec} \mathcal{L}_{rec},
\end{equation}
where $\mathcal{L}_{\textrm{ST-ED}}$ represents the losses used in ST-ED~\cite{sted} excluding the original $\ell1$ loss.

\textbf{Implementation Details.} 
We follow the same setting as the original ST-ED~\cite{sted} for the normalization settings and training hyperparameters including the learning rate.
The weight for the mixed reconstruction loss is $\alpha = 0.84$~\cite{Nitzan2020FaceID}.
We emperically set $\lambda_{id}=2$ and $ \lambda_{rec}=200$.
The AdaFace encoder is a ResNet-50 model~\cite{resnet_He_cvpr16}, and we intentionally freeze it and only train the other parts DenseNet encoder and decoder to avoid overfitting on the training subjects.


\section{Experiments}

\begin{table*}[t]
\centering
\caption{ Redirection performance on XGaze Test. The first column is the training data and the second column is the redirection model. 
$\dagger$: GC short for GazeCapture} 
\label{tab:main_eval_xgaze_test}
\scalebox{1.1}{
\begin{tabular}{ll|ccccc}\hline
    Training Data &  Model & Head $\downarrow$ & Gaze $\downarrow$ & LPIPS $\downarrow$ & FID $\downarrow$  & Sim. ($\uparrow$)\\\hline
    XGaze-SH & ST-ED &  6.13  & 11.56  &  0.203  &  48.01  & 0.267 \\\hline
    \multirow{2}{*}{XGaze}   & ST-ED &2.92 & 7.93  & 0.173  & 42.88   & 0.331 \\
       & \adasted &  2.94 & 7.56  & 0.163  & 36.30  & 0.351 \\
    \Xhline{2\arrayrulewidth}
    GC$~\dagger$  & ST-ED &	22.04 &   33.83 &  0.380  &  137.97  &  0.118\\
    \hline
    \multirow{2}{*}{GC + Aug-60H}   & ST-ED &	9.27 &  \textbf{18.89}  &  0.324 &   116.10  &    0.142 \\
  & \adasted  &   \textbf{8.08} & 19.03  &  \textbf{0.290}  &  \textbf{109.08}  & \textbf{0.210} \\
    \hline
\end{tabular}
}
\end{table*}

\subsection{Experimental Settings}\label{sec:evaluation_settings}

For evaluation setting, previous work calculates the redirection error of a model when redirecting to a target image (\textit{redirect-to-image}) by using ground truth paired images~\cite{cuda_ghr_jindal2021, sted, he2019GAN_eye, ganin2016deepwarp}.
However, this evaluation method is limited by the angle range of the test data, that is, it cannot evaluate the redirection error under large target angles if the test dataset has a limited range. 
Therefore, we also evaluate the redåirection error when directly redirecting to a target angle (\textit{redirect-to-angle}).
Specifically, we randomly sample 10 new directions from a uniform distribution with a radius of 60\textdegree.
We compute the rotation matrix based on the source and target head pose and apply the rotation matrix on both the head and gaze embeddings.
We sample 20 samples from each subject of GazeCapture as the source image since GazeCapture's images are almost frontal.

To obtain the head pose and gaze direction of the generated image, we use a ResNet-18~\cite{resnet_He_cvpr16} estimator.
The estimator is trained on the corresponding dataset when evaluating the model on each dataset such that it can predict accurate head pose and gaze.
The average estimation errors of head/gaze for GazeCapture and ETH-XGaze are 0.88/1.37 degrees and 0.61/0.83 degrees, respectively, indicating that the estimator is reliable.
For pre-processing, we adopt the data normalization~\cite{zhang18_etra}, which is commonly used in gaze-related tasks~\cite{sted, ETHXGaze_Zhang2020}.

\subsection{Evaluation Method}
\textbf{Metrics}
We compute the \textbf{redirection angular error} between the ground-truth target direction and the estimated direction of the generated image.
Both the head redirection error and the gaze redirection error are evaluated.
We adopt \textbf{LPIPS}~\cite{lpips_zhang2018perceptual} to evaluate the general similarity of the image between the generated image and the target image. 
Besides the above metrics used in previous works, we further evaluate the image quality of generation by \textbf{Fréchet Inception Distance} (FID)~\cite{fid_heusel2017gans, fid_Seitzer2020}, which has been widely used to evaluate the visual similarity of two groups of images that is close to human perception.
In addition, to evaluate the identity preservation, we also compute the \textbf{identity similarity} between the target image and generated image using AdaFace~\cite{kim2022adaface}. 
Notice this evaluator is different from the one used for training.

\noindent\textbf{Baseline methods}
Besides, since we focus on data creation, we only compare with the SOTA generative model \textbf{ST-ED}~\cite{sted}, which hugely outperforms most previous methods~\cite{he2019GAN_eye, choi2018stargan} and is still more indicative and suggestive than the latest neural-radiance-fields-based (NeRF) redirection methods~\cite{yin2022nerf, ruzzi2022gazenerf}.

\begin{table}[t]
\caption{The average angular error (degree) of redirection to target direction.
$\dagger$: GC short for GazeCapture
}
\label{tab:gc-red-both}
\begin{center}
\scalebox{1.05}{
    \begin{tabular}{ll|cc}\hline
    Training Data & Model &  Head Error  $\downarrow$ & Gaze Error $\downarrow$  \\ \hline
    GC~$\dagger$  & ST-ED &	4.96 &  16.42 \\\hline
   \multirow{2}{*}{GC + Aug-60H}  & ST-ED &  3.63 & 12.37  \\
     & \adasted & \textbf{3.67} &  \textbf{11.09} \\\hline
    \end{tabular}\hfil
    }
\end{center}
\end{table}

\begin{figure}[t]
\centering
    \includegraphics[width=0.99\linewidth]{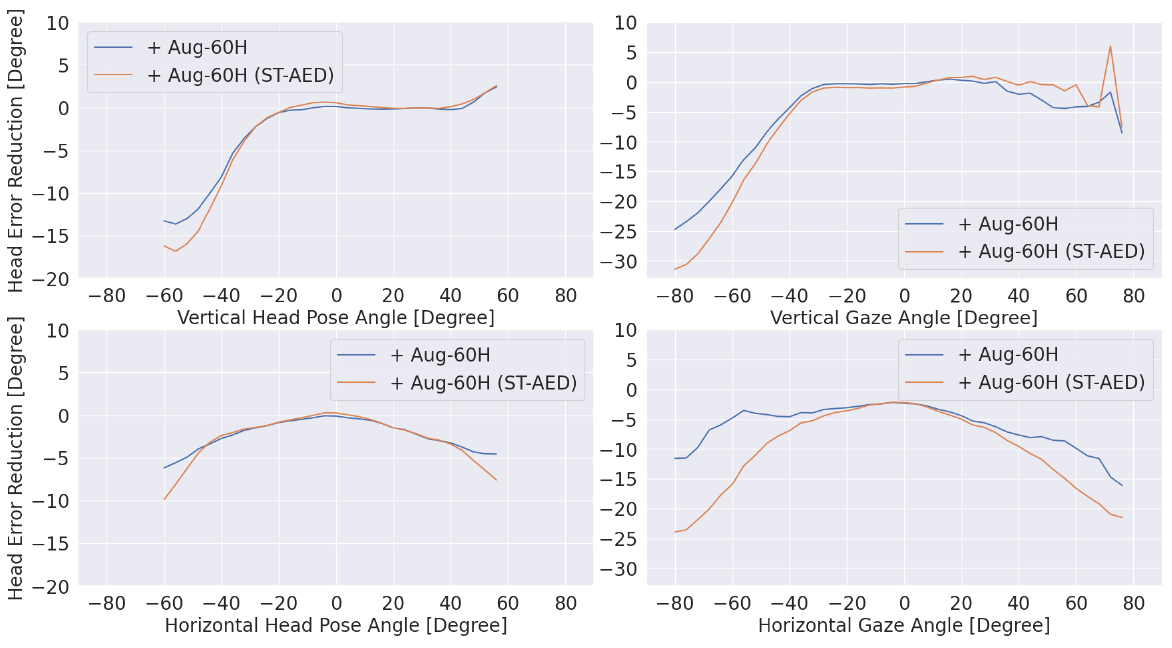}
    \caption{The error reduction (degree) distribution of head pose (left side) and gaze (right side) by different target angle for each setting by the proposed augmentation Aug-60H and~\adasted.
    The error reduction of the proposal is more obvious under larger target angles.
    }
    \label{fig:err_diff_both-gc}
\end{figure}

\subsection{Experiments Result}
\subsection*{Redirection} 

For the \textit{redirect-to-image} result shown in Table~\ref{tab:main_eval_xgaze_test}, we use XGaze Test as the testing data.
Table~\ref{tab:gc-red-both} shows the result of the \textit{redirect-to-angle} evaluation.
As a reference, we first compare the results of using XGaze and its subset XGaze-SH as training data, the angle distribution and qualitative results of which were previously shown in Fig.~\ref{fig:teaser}.
The subset XGaze-SH with a small angle range showed worse performance than XGaze on all metrics in Table~\ref{tab:main_eval_xgaze_test}, showing the significant influence of the limited angle range of the training data.

\begin{figure}[t]
\centering
    \includegraphics[width=0.95\linewidth]{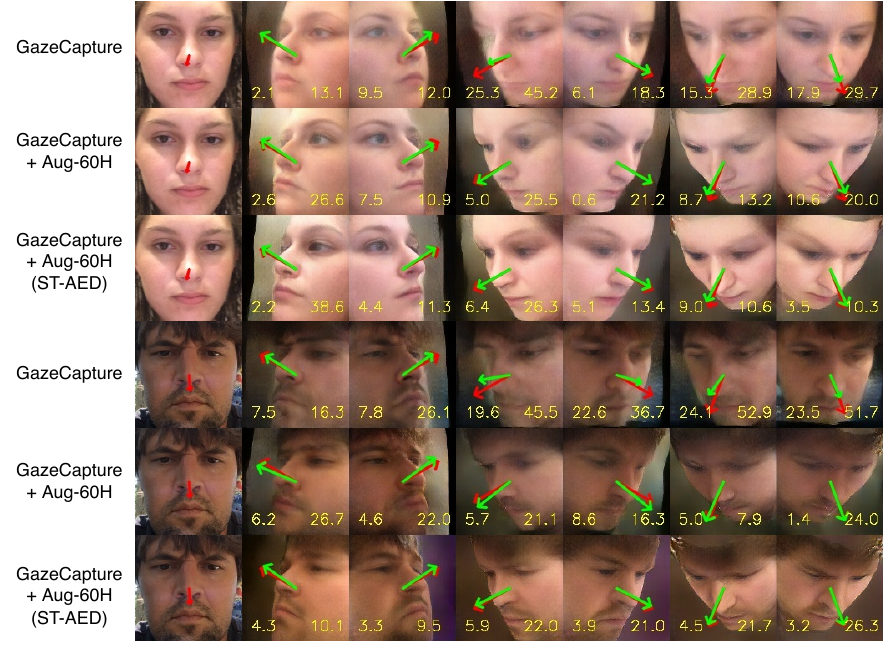}
    \caption{ Examples of redirecting images to the target direction.
    The values at the bottom are the redirection error of the head pose (left) and gaze (right), and the arrows are the visualization of the ground truth and the estimated head pose.}
    \label{fig:red_samples}
\end{figure}

\textbf{Angle Extension.}
For GazeCapture as training data (GC), as shown in the fourth row of Table~\ref{tab:main_eval_xgaze_test}, 
it showed very large redirection errors since XGaze Test has a much larger direction range than GazeCapture.
Similar results can also be observed from the first row of Table~\ref{tab:gc-red-both}.
The proposed augmentation-based training data (GC+Aug-60H), as expected, showed a significant reduction in the head redirection error (by about 13\textdegree) and gaze redirection error (by about 15\textdegree) because it extends the angle range of training data to be similar to the XGaze Test.
In Table~\ref{tab:gc-red-both}, since the process of creating augmentation data as described in Section~\ref{aug:method} is similar to the \textit{redirect-to-angle} pattern, the data augmentation shows effective reduction on both head and gaze error.
For the further reduction of~\adasted~in the last row of both Table~\ref{tab:main_eval_xgaze_test} and Table~\ref{tab:gc-red-both}, it is reasonable to speculate that it is partially due to the improved image quality, making the estimator easier to predict the direction.

Fig.~\ref{fig:err_diff_both-gc} further shows the redirection error reduction distribution according to its target angle. 
The reduction is based on the GazeCapture-only (GC) baseline, which corresponds to the first row in Table~\ref{tab:gc-red-both}. 
The negative values mean that the error is reduced from the baseline.
We can observe that the error reduction of the proposal is more obvious and effective under large target angles.
From the examples of redirected images in Fig.~\ref{fig:red_samples}, we can observe that the source image can be redirected to a larger angle through the proposed data augmentation.

\textbf{Image Quality.} 
First, by comparing GC and GC+Aug-60H (ST-ED) in Table~\ref{tab:main_eval_xgaze_test}, the proposed augmentation (Aug-60H) also improved the image quality on metrics of LPIPS, FID, and identity similarity, because it can generate better head redirection.
Moreover, the proposed~\adasted~further improved the identity preservation and image quality, as can be seen from the last row of Table~\ref{tab:main_eval_xgaze_test}.
From the qualitative results in Fig.~\ref{fig:err_diff_both-gc}, we can observe that the images generated by~\adasted~(the third and sixth rows) show more photo-realistic appearance than those by the ST-ED baseline.


\subsection*{Ablation Studies}

\begin{figure}[t]
\centering
    \includegraphics[width=0.99\linewidth]{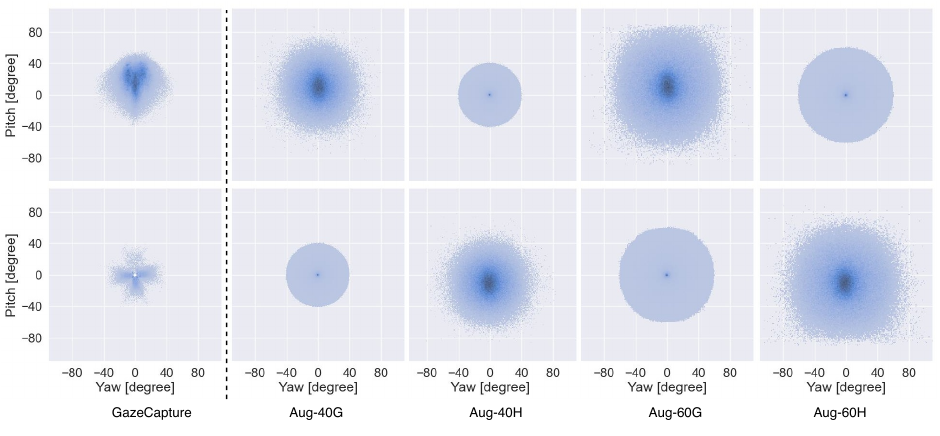}
    \caption{The label distribution of GazeCapture and the augmentation datasets. 
    The top row is the head pose distribution and the bottom row is the gaze distribution.}
    \label{fig:gc_aug_distribution}
\end{figure}

\begin{figure}[t]
\centering
    \includegraphics[width=0.9\linewidth]{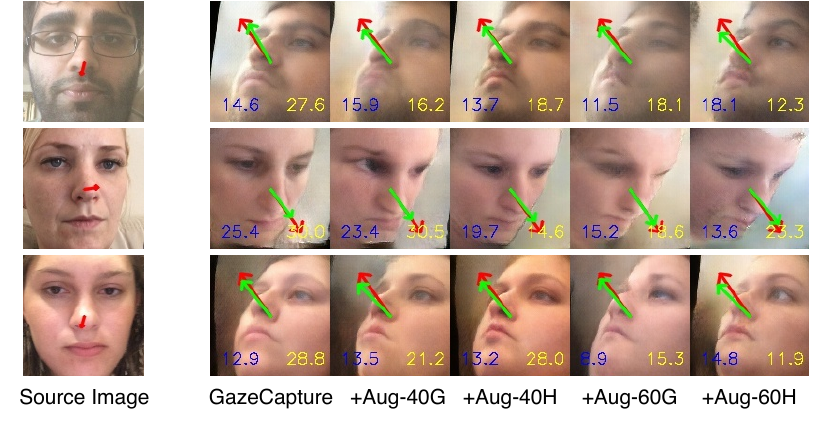}
    \caption{ Examples of \textit{redirect both}.
        The blue and yellow numbers correspond to the redirection error of head and gaze, respectively.
        The red and yellow arrows correspond to the ground-truth gaze direction and the estimated gaze direction, respectively.}
    \label{fig:abla1_both}
\end{figure}

\begin{figure}[t]
\centering
    \includegraphics[width=0.9\linewidth]{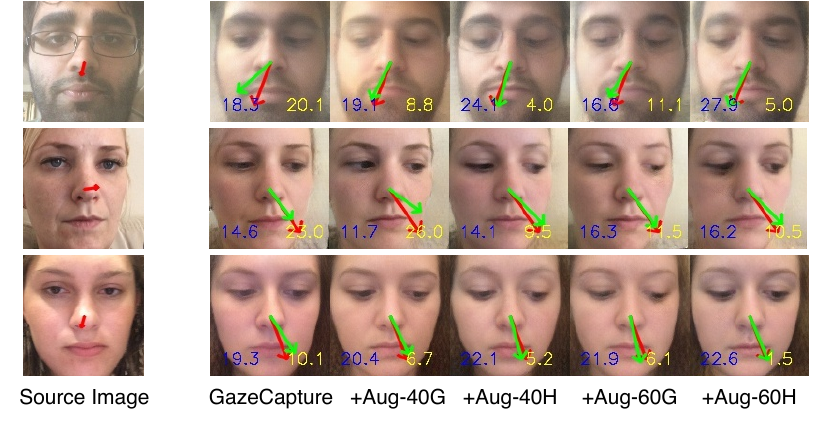}
    \caption{Examples of \textit{redirect gaze}.
    }
    \label{fig:abla1_gaze}
\end{figure}

\begin{figure}[t]
\centering
    \includegraphics[width=0.9\linewidth]{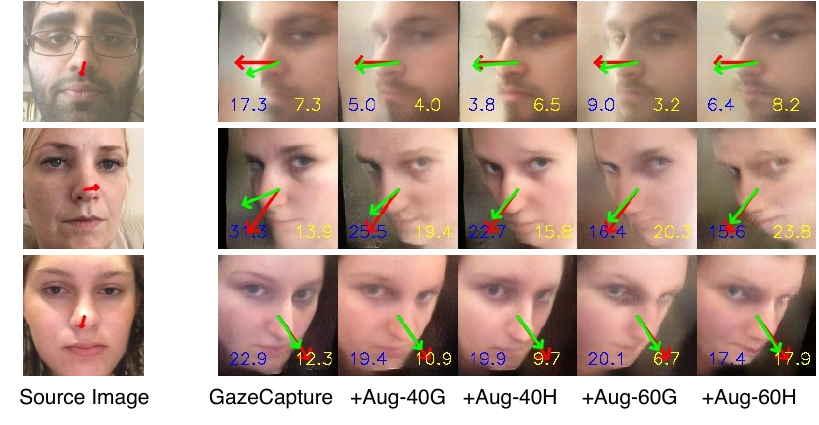}
    \caption{Examples of \textit{redirect head}.
    }
    \label{fig:abla1_head}
\end{figure}

\textbf{Data creation patterns.}
Besides the \textit{head-based} sampling, we also create augåmentation datasets with \textit{gaze-based} sampling, where the rotation matrix is computed based on the source and target gaze direction.
In addition to the 60-degree sampling, to further evaluate the impact of the angle range of the augmentation data, we also sample the target direction from a 40-degree circle-shaped uniform distribution.
Besides augmenting based on the head pose, we also generate samples based on the target gaze. 
As a result, we've produced four augmentation datasets for comparison, named Aug-40G, Aug-40H, Aug-60G, and Aug-60H, whose distributions are shown in Fig.~\ref{fig:gc_aug_distribution}, where G and H represent head-pose-based sampling and gaze-based sampling, respectively.

We compare these datasets by the \textit{redirect-to-angle} evaluation.
In addition, we consider three redirection patterns.
1) Redirect both: we compute the rotation matrix based on source and target head pose, and apply the rotation matrix on head and gaze embeddings at the same time;
2) Gaze only: we compute the rotation matrix based on source and target gaze, and apply the rotation matrix only on gaze embeddings, which means fixing the original head pose;
3) Head only: we compute the rotation matrix based on the source and target head pose, and apply the rotation matrix only on head embeddings, which means fixing the original gaze direction.

\begin{table*}[t]
\centering
\caption{The redirection error in three patterns for GazeCapture and GazeCapture with different augmentation datasets.}
\label{tab:gaze-red-all}
\scalebox{1.1}{
    \begin{tabular}{l|cc|cc|cc}\hline 
    & \multicolumn{2}{c|}{Both} & \multicolumn{2}{c|}{Gaze Only} & \multicolumn{2}{c}{Head Only}\\ 
    & Head $\downarrow$ & Gaze $\downarrow$  & Head  $\downarrow$ & Gaze $\downarrow$  & Head $\downarrow$ & Gaze $\downarrow$  \\\hline
    GazeCapture  & 5.85  & 17.15  & 1.62   & 12.59   &  5.81  &  3.81 \\\hline
    + Aug-40G  &  4.62 & 14.67  &  1.66  &  11.53  &  4.28 & 4.09 \\
    + Aug-40H  & 4.28  & 12.52  &  1.63  &  11.22  &  4.32 &  4.01 \\
    + Aug-60G  & 4.17  & 12.63  &  \textbf{1.46}  &  11.16  & \textbf{4.16}  & 3.83 \\
    + Aug-60H  & \textbf{3.67}  & \textbf{11.09}  &  1.57  &  \textbf{10.76}  &  4.24 &  \textbf{3.81}\\\hline
    \end{tabular}\hfil
    }
\end{table*}


\begin{figure}[t]
\centering
    \includegraphics[width=0.9\linewidth]{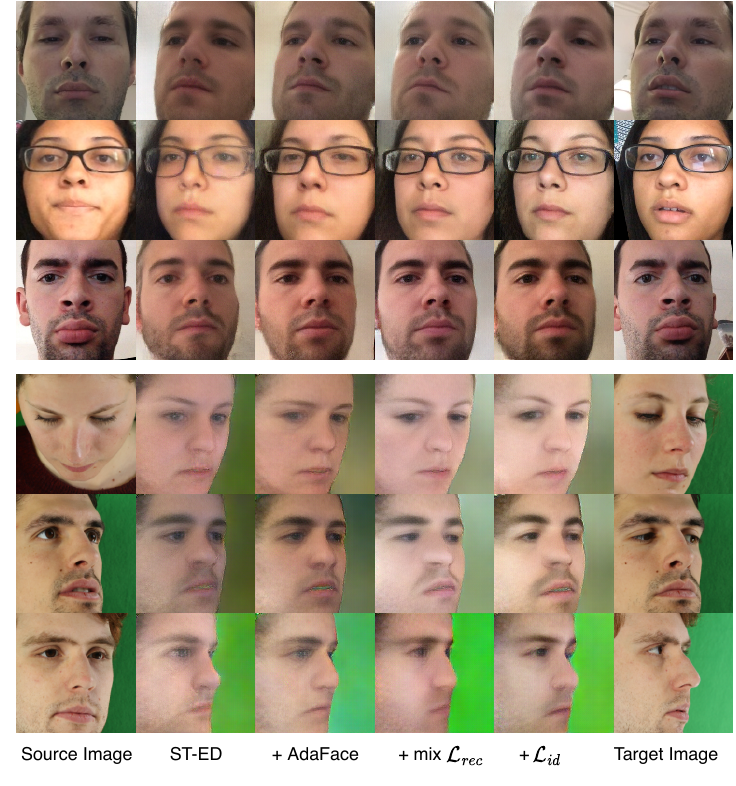}
    \caption{ Examples of redirected images that can show the difference of face identity and image quality. 
    Each column corresponds to the row in Table~\ref{tab:abla2_eval_gc_test} and Table~\ref{tab:abla2_eval_xgaze_test}.
    The first three rows are from GazeCapture Test, and the last three rows are from XGaze Test.}
    \label{fig:abla2_samples}
\end{figure}

\begin{table*}[t]
\centering
\caption{The ablation study of \textit{redirect-to-image} evaluation for GazeCapture Test.
The second to fourth columns are the components in the proposed~\adasted.
The last row with all components corresponds to the proposal~\adasted.}
\label{tab:abla2_eval_gc_test}
\scalebox{0.99}{
\begin{tabular}{c|ccc|ccccc}\hline
    Training Data & AdaFace & mix $\mathcal{L}_{rec}$  &$\mathcal{L}_{id}$  &  Head $\downarrow$  & Gaze $\downarrow$ & LPIPS $\downarrow$ & FID $\downarrow$  & Sim. $\uparrow$ \\\hline
    \multirow{4}{*}{ \makecell{GC  \\ + Aug-60H} } &  &  &    &  1.26 &  3.86  &  0.191 &  41.76 &  0.472  \\
   & \checkmark   &  &     &   1.14 &  3.78  &   0.176  &  \textbf{36.30}  &  0.511  \\
   & \checkmark   & \checkmark  &  &  \textbf{1.13} &  3.77  &   0.175  &  38.91  &  0.534 \\
   & \checkmark  &  \checkmark &  \checkmark  &  1.19 &  \textbf{3.76} & \textbf{0.175} &  39.24  & \textbf{0.535} \\\hline
\end{tabular}
}
\end{table*}

\begin{table*}[t!]
\centering
\caption{The ablation study of \textit{redirect-to-image} evaluation for XGaze Test.
The second to fourth columns are the components in the proposed~\adasted.
The last row with all components corresponds to the proposal~\adasted.}
\label{tab:abla2_eval_xgaze_test}
\scalebox{0.99}{
\begin{tabular}{c|ccc|ccccc}\hline
    Training Data & AdaFace & mix $\mathcal{L}_{rec}$  &$\mathcal{L}_{id}$  &  Head $\downarrow$  & Gaze $\downarrow$ & LPIPS $\downarrow$ & FID $\downarrow$  & Sim. $\uparrow$ \\\hline
    \multirow{4}{*}{ \makecell{GC  \\ + Aug-60H} }  &  &  &      & 9.27	& 18.89 & 0.324 & 116.10 & 0.142 \\
    &  \checkmark &  &    & 9.76 & 20.63 & 0.317 & 117.02 & 0.160\\
    & \checkmark  &\checkmark  &     &	8.18 & \textbf{18.64} & 0.290  & \textbf{105.11} & 0.196 \\
    & \checkmark & \checkmark & \checkmark    &	\textbf{8.08} & 19.03 & \textbf{0.290} & 109.08  & \textbf{0.210} \\\hline
\end{tabular}
}
\end{table*}

The average redirection errors of the three patterns are shown in Table~\ref{tab:gaze-red-all}.
While all augmentation datasets showed a reduction in both head error and gaze error,  60-degree augmentation shows a better performance, which is intuitively reasonable.
However, this range should not be overly enlarged since the facial region may be totally invisible.
Since the process of creating Aug-60H is similar to the \textit{redirect both} pattern, it shows the best performance as in the second to third columns of Table~\ref{tab:gaze-red-all}. 
For \textit{gaze only} and \textit{head only}, the Aug-60G showed the best head error while the Aug-60H showed the best gaze error.
We speculate the reason to be that the Aug-60G has a larger head pose range while the Aug-60H has a larger gaze range, as can be observed in Fig.~\ref{fig:gc_aug_distribution}.
Finally, we show qualitative results of the three patterns in Fig~\ref{fig:abla1_both}, Fig~\ref{fig:abla1_gaze}, and Fig~\ref{fig:abla1_head}, respectively.
Further video samples are also included in the supplementary materials.

\textbf{\adasted~Components.}
We conduct an ablation study using the \textit{redirect-to-image} evaluation to compare the performance of individual components of the model. 
We use the proposed augmentation-based training data (GC+Aug-60H) as the training data for all settings.
We use two datasets, GazeCapture Test and XGaze Test, to evaluate the model's effectiveness in terms of image quality and identity preservation. 
From the results shown in Table~\ref{tab:abla2_eval_gc_test} and~\ref{tab:abla2_eval_xgaze_test}, 
the proposed AdaEncoder, mixed reconstruction loss, and identity loss all contribute to improving the performance, especially identity similarity. 
Qualitative results are also shown in Fig.~\ref{fig:abla2_samples}.
In the GazeCapture Test, the~\adasted~effectively generates detailed and more target-like images. 
In the XGaze Test, we observe that simply training the baseline ST-ED model with synthetic data results in low-quality images, as shown in the second column of Fig.~\ref{fig:abla2_samples}. 
While the proposed~\adasted~showed a tendency to generate images with less blurred eye regions and better facial shapes under large angles.


\section{Conclusion and Future Work}

\textbf{Conclusion.} For learning-based gaze and head redirection, we proposed using 3D face reconstruction-based synthetic data to tackle the limited angle range of existing training datasets.
Experiments showed effectiveness in terms of redirection accuracy.
To better train with the synthetic images, we further proposed a corresponding gaze and head redirection framework~\adasted, which can generate face images with satisfying quality even under large target angles.

\noindent\textbf{Future Work.} 
Although the main contribution of this work is to create effective augmentation data for extending the redirection angle range and improving the redirection accuracy, we acknowledge that there is still room for improvement in the image quality.
This issue is largely tied to the quality of the original training data, with datasets such as GazeCapture providing lower quality than the ETH-XGaze dataset (see Fig.~\ref{fig:teaser}).
Therefore, the image quality and corresponding identity problem is indeed a challenging issue in future work,  with one potential solution being to further investigate transfer learning among datasets without compromising the redirection accuracy.

\bibliographystyle{IEEEtran}
\bibliography{reference}



\end{document}